\pdfoutput=1

\documentclass[11pt]{article}

\usepackage{acl}

\usepackage{times}
\usepackage{latexsym}

\usepackage[T1]{fontenc}

\usepackage[utf8]{inputenc}

\usepackage{microtype}

\usepackage{inconsolata}

\usepackage{graphicx}
\usepackage{wrapfig}
\usepackage{amssymb}
\usepackage{color,xcolor,colortbl}
\usepackage{enumitem}
\usepackage{soul}
\usepackage{amsmath}
\usepackage{url}
\usepackage{algorithm}
\usepackage[noend]{algpseudocode}
\usepackage{booktabs}
\usepackage{bm,bbm}
\usepackage{mathtools}
\usepackage{array}
\usepackage{multirow}
\usepackage{subcaption}
\usepackage{pbox}
\usepackage{pifont}
\usepackage{arydshln}
\usepackage{dblfloatfix}
\usepackage{relsize}
\usepackage{setspace}
\usepackage[scaled=.8]{beramono}

\definecolor{darkblue}{rgb}{0.0, 0.0, 0.55}

\definecolor{gray}{HTML}{708090}
\definecolor{gem}{HTML}{80A8EE}
\definecolor{claude}{HTML}{cc785c}
\definecolor{darkgreen}{HTML}{0FA37F}
\definecolor{lightgreen}{HTML}{51DA4C}
\definecolor{mistral}{HTML}{FE4A00}
\definecolor{llama}{HTML}{0068E8}

\title{\texttt{SportsMetrics}: Blending Text and Numerical Data to Understand Information Fusion in LLMs}

\author{Yebowen Hu,$^\dagger$ Kaiqiang Song,$^\ddagger$ Sangwoo Cho,$^\ddagger$ Xiaoyang Wang,$^\ddagger$\\ 
\textbf{Hassan Foroosh,$^\dagger$ Dong Yu,$^\ddagger$ Fei Liu$^\S$}\\[0.5em]
$^\dagger$University of Central Florida \, 
$^\ddagger$Tencent AI Lab, Bellevue, WA \, 
$^\S$Emory University\\
\texttt{\{yebowen.hu, hassan.foroosh\}@ucf.edu}\\
\texttt{\quad \{riversong, swcho, shawnxywang, dyu\}@global.tencent.com fei.liu@emory.edu}
}

\begin{document}
\maketitle
\begin{abstract}

Large language models hold significant potential for integrating various data types, such as text documents and database records, for advanced analytics. However, blending text and numerical data presents substantial challenges. LLMs need to process and cross-reference entities and numbers, handle data inconsistencies and redundancies, and develop planning capabilities such as building a working memory for managing complex data queries. In this paper, we introduce four novel tasks centered around sports data analytics to evaluate the numerical reasoning and information fusion capabilities of LLMs. These tasks involve providing LLMs with detailed, play-by-play sports game descriptions, then challenging them with adversarial scenarios such as new game rules, longer durations, scrambled narratives, and analyzing key statistics in game summaries. We conduct extensive experiments on NBA and NFL games to assess the performance of LLMs on these tasks. Our benchmark, \texttt{SportsMetrics}, introduces a new mechanism for assessing LLMs' numerical reasoning and fusion skills.

\end{abstract}

\section{Introduction}
\label{sec:intro}

\begin{figure}
\centering
\includegraphics[width=2.78in]{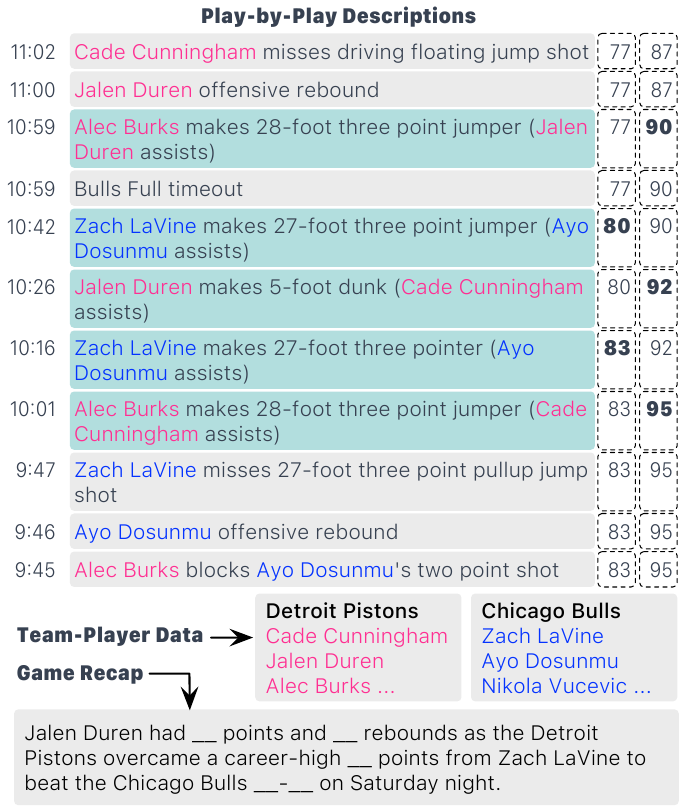}
\vspace{-0.05in}
\caption{Play-by-plays of an NBA game. We include timestamps, player actions, team affiliations and a game recap. Total points for both teams are indicated in dotted circles and are withheld from LLMs.
}
\label{fig:example-data}
\vspace{-0.2in}
\end{figure}

Large language models (LLMs) are more powerful than ever. OpenAI's GPT-4 Turbo~\shortcite{gpt4-turbo} features a 128k context window, allowing it to process over 300 pages of text in a single prompt. Claude v2.1~\shortcite{claude-2.1} steps it up with a 200k token window, equivalent to roughly 150,000 words or more than 500 pages. Mistral AI~\shortcite{mixtral-8x7b} has created a sparse mixture of experts model capable of processing up to 32k tokens. The developments suggest language models can now engage with vast amounts of text content and data, opening doors to exciting new applications in various domains.

One of the most promising uses of LLMs is in handling a combination of unstructured texts and structured data. For example, determining if a patient can be discharged from the hospital may involve reviewing doctor notes, radiology and pathology reports, lab results, and other records that blend text and structured data~\cite{adams-etal-2021-whats,bardhan-etal-2022-drugehrqa,cai-etal-2022-generation,cai2023paniniqa,vanveen2023clinical,ben-abacha-etal-2023-empirical}; LLM Assistants for online shopping need to process product catalogs, sales transactions, and customer queries~\cite{brynjolfsson2023generative,wendys-chatbot}. Yet, summarizing key details from a mix of unstructured and structured sources remains a considerable challenge. An LLM must navigate text descriptions, link entities, aggregate numbers, handle discrepancies, and beyond.

\begin{figure*}
\centering
\includegraphics[width=6.3in]{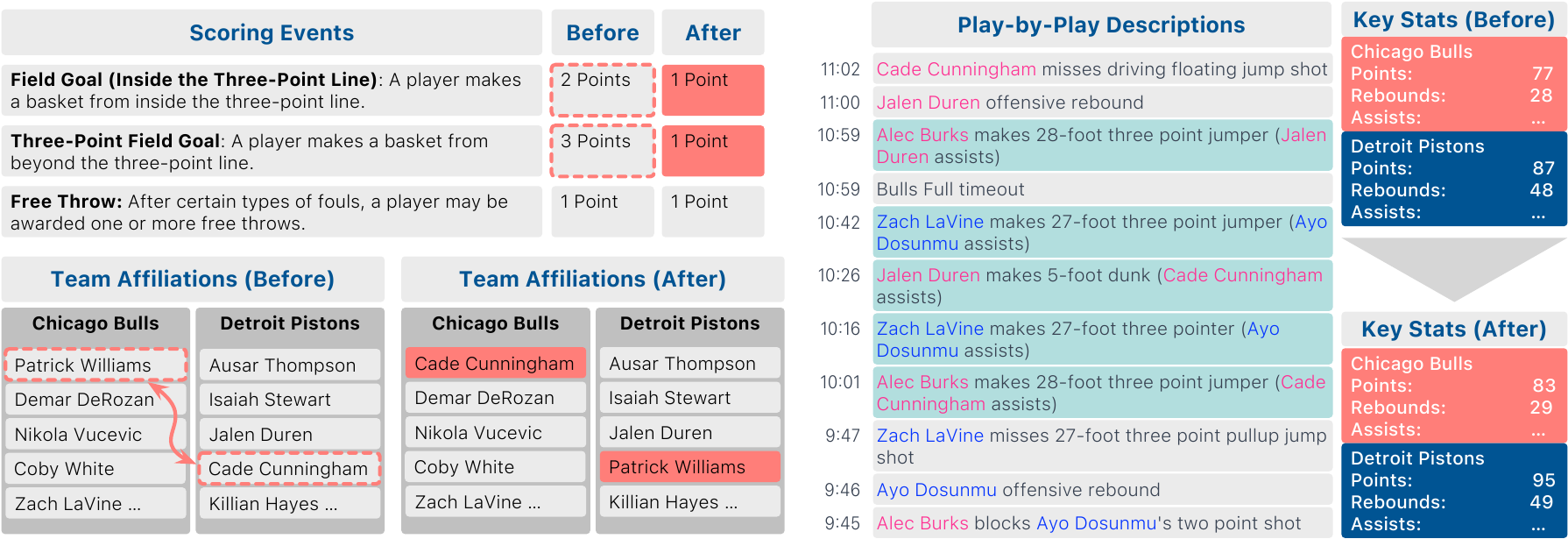}
\vspace{-0.15in}
\caption{
(\textsc{Top Left}) We examine the impact of changing game rules on final scores. For basketball, scoring events such as free throws, three-pointers, field goals, vary from 1 to 3 points. We ask LLMs to maintain these scoring events but under a new rule where each is worth only 1 point. (\textsc{Bottom Left}) We randomly swapped player team affiliations in the table without altering the game's play-by-play records. (\textsc{Right}) LLMs are provided with detailed play-by-play descriptions of a sports game and player team affiliations. Their job is to use this information to update key game statistics in a JSON format.
}
\label{fig:tasks}
\vspace{-0.1in}
\end{figure*}

Information fusion focuses on synthesizing information from multiple textual sources to derive meaningful conclusions~\cite{barzilay-etal-1999-information}. Current approaches involve summarizing multiple text documents, providing concise answers to user queries, and integrating summarization with natural language inference to deduce information~\cite{bhaskar-etal-2023-prompted,caciularu-etal-2023-peek,sprague-etal-2022-natural,bostrom-etal-2022-natural}. The output is often a short text summary, the quality of which is difficult to evaluate~\cite{deutsch-etal-2021-statistical}. In contrast, our approach emphasizes the numerical aspect of information fusion~\cite{geva-etal-2020-injecting,zhu2021tatqa,zhao2023docmatheval,reddy2024docfinqa}. We enable the LLM to navigate through lengthy texts, gather crucial statistics, and develop a working memory to manage complex data queries.

\begin{figure*}
\centering
\includegraphics[width=6.3in]{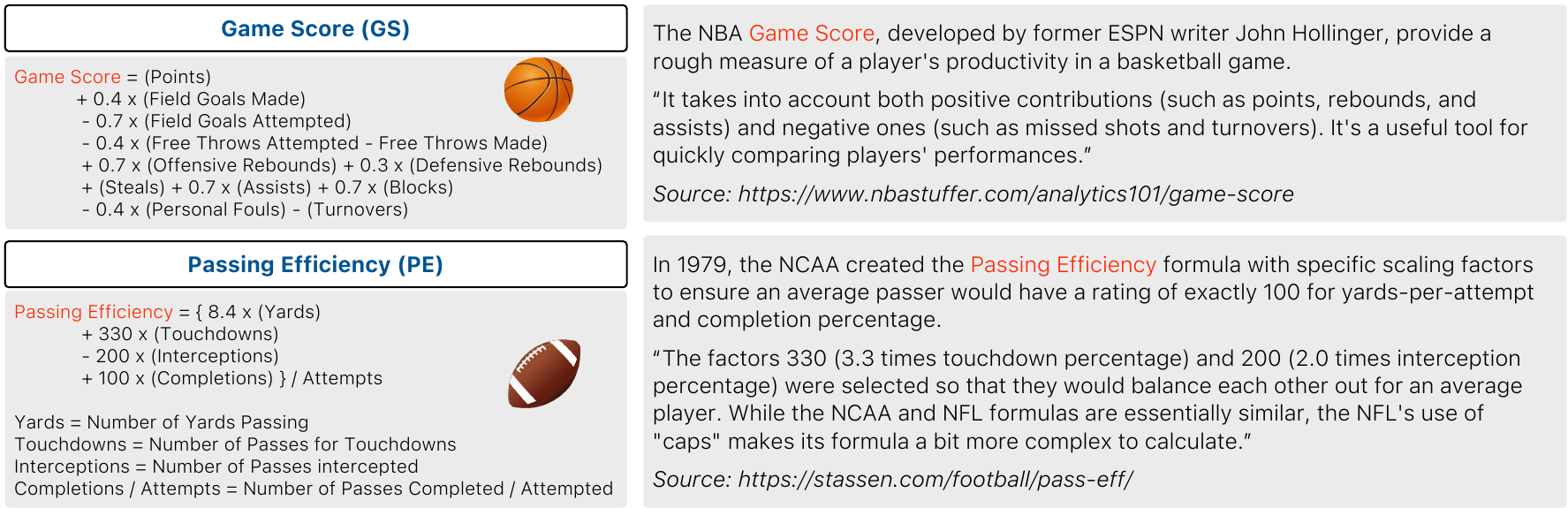}
\caption{We adopt the NBA's \emph{Game Score}, originally designed for player evaluation, to measure a team's overall efficiency. For American football, we apply NCAA's \emph{{Passing Efficiency}} formula.}
\label{fig:game-scores}
\vspace{-0.1in}
\end{figure*}

We introduce \texttt{SportsMetrics}, a benchmark designed to assess LLMs' abilities in numerical reasoning and data fusion. This benchmark provides LLMs with detailed, play-by-play descriptions of sports games, including timestamps, player actions, and team affiliations, as illustrated in Figure~\ref{fig:example-data}. We focus on four novel tasks to evaluate LLMs in adversarial scenarios: (a) \emph{adapting to new game rules}, (b) \emph{handling lengthy game descriptions}, (c) \emph{managing scrambled game narratives}, and (d) \emph{analyzing critical statistics in game summaries}. E.g., an LLM might be asked to complete a basketball game recap by inserting missing key statistics, which requires the development of a working memory for game stats and reasoning skills.

Our \texttt{SportsMetrics} benchmark presents three main benefits. First, it leverages sports data, including team-player affiliations and play-by-play details; they are dynamic narratives that LLMs cannot easily memorize. Second, it allows us to evaluate LLMs' ability to track key statistics such as team points, assists, blocks, steals, and more, while also offering an overall game efficiency score for direct LLM comparison. Lastly, its use of widely understood sports terminology makes it more accessible to researchers than specialized medical language, making it an ideal benchmarking tool. While our current focus is on English, \texttt{SportsMetrics} also holds promise for multilingual applications.

\section{Related Work}
\label{sec:related}

There is a growing need for a benchmark to evaluate LLMs' \emph{information fusion} capabilities, which offers clear, quantitative scores for comparing various LLMs. For example, Chatbot Arena~\cite{zheng2023judging} utilizes Elo ratings~\cite{boubdir2023elo}, MT-Bench comprises of 80 multi-turn questions, and MMLU focuses on a model's multitask accuracy across 57 tasks~\cite{hendrycks2021measuring}. Multi-document summarization offers a promising benchmark~\cite{lebanoff-etal-2021-modeling,huang-etal-2021-efficient,wang-etal-2022-squality,xu-etal-2023-lmgqs}. However, developing a summary scoring system poses challenges due to variables such as summary length, content coverage, and faithfulness~\cite{cao-etal-2022-hallucinated,liu2023benchmarking,krishna-etal-2023-longeval,hu-etal-2023-decipherpref,li-etal-2023-defining,xu2024identifying,joseph2024factpico}. 
Sports data, which combines static knowledge with player dynamics, presents an untapped opportunity for benchmarking LLMs.

Combining information from a blend of textual and numerical records poses a significant challenge. In traditional multi-document summarization, the system creates a concise summary from a set of topically related documents. Giorgi et al.~\cite{giorgi-etal-2023-open} show that this task remains difficult in an ``open-domain'' setting, where the document set is generated by a retriever and may include irrelevant information. With the growing popularity of retrieval-augmented generation (RAG)~\cite{karpukhin-etal-2020-dense,liu-etal-2022-challenges}, there is an increasing need to accurately fuse information from various sources. We explore information fusion by examining how LLMs cross-reference players and actions, and aggregate data across play-by-play descriptions to compile key game statistics.

Our work relates to numerical reasoning, which uses arithmetic reasoning to tackle mathematical word problems. Prior datasets in this area include MathQA~\cite{amini2019mathqa}, GSM8k~\cite{cobbe2021training}, SVAMP~\cite{patel2021nlp}, TAT-QA~\cite{zhu2021tatqa}, FinQA~\cite{chen2022finqa}, MATH~\cite{liu2023improving}, DocMath-Eval~\cite{zhao2023docmatheval}, TABMWP~\cite{lu2023dynamic} and more, many allowing models to generate answer explanations. The problems typically have brief descriptions, with the challenge lying in creating an expression tree and applying arithmetic knowledge. In contrast, our approach focuses on assessing LLMs' ability to track key statistics across extremely long contexts.

Sports data has been utilized in various natural language tasks, including data-to-text generation for sports games~\cite{lareau-etal-2011-detecting,zhang-etal-2016-towards,wiseman-etal-2017-challenges,van-der-lee-etal-2017-pass,puduppully-etal-2019-data}, real-time game summarization from live commentaries~\cite{edouard-etal-2017-youll,huang-etal-2020-generating}; and other aspects such as sports commentator bias~\cite{merullo-etal-2019-investigating}. Beyond sports, there's significant interest in annotating and analyzing large-scale game-related corpora, such as reviews and gameplay logs, and summarizing gameplay commentaries~\cite{gamesandnlp-2020-games,kicikoglu-etal-2020-aggregation,gu-etal-2022-revisiting,furman-etal-2022-sequence}. We anticipate that insights from our \texttt{SportsMetrics} benchmark will benefit these areas, enhancing our understanding of game narratives and player dynamics.

\begin{figure*}
\centering
\includegraphics[width=6.3in]{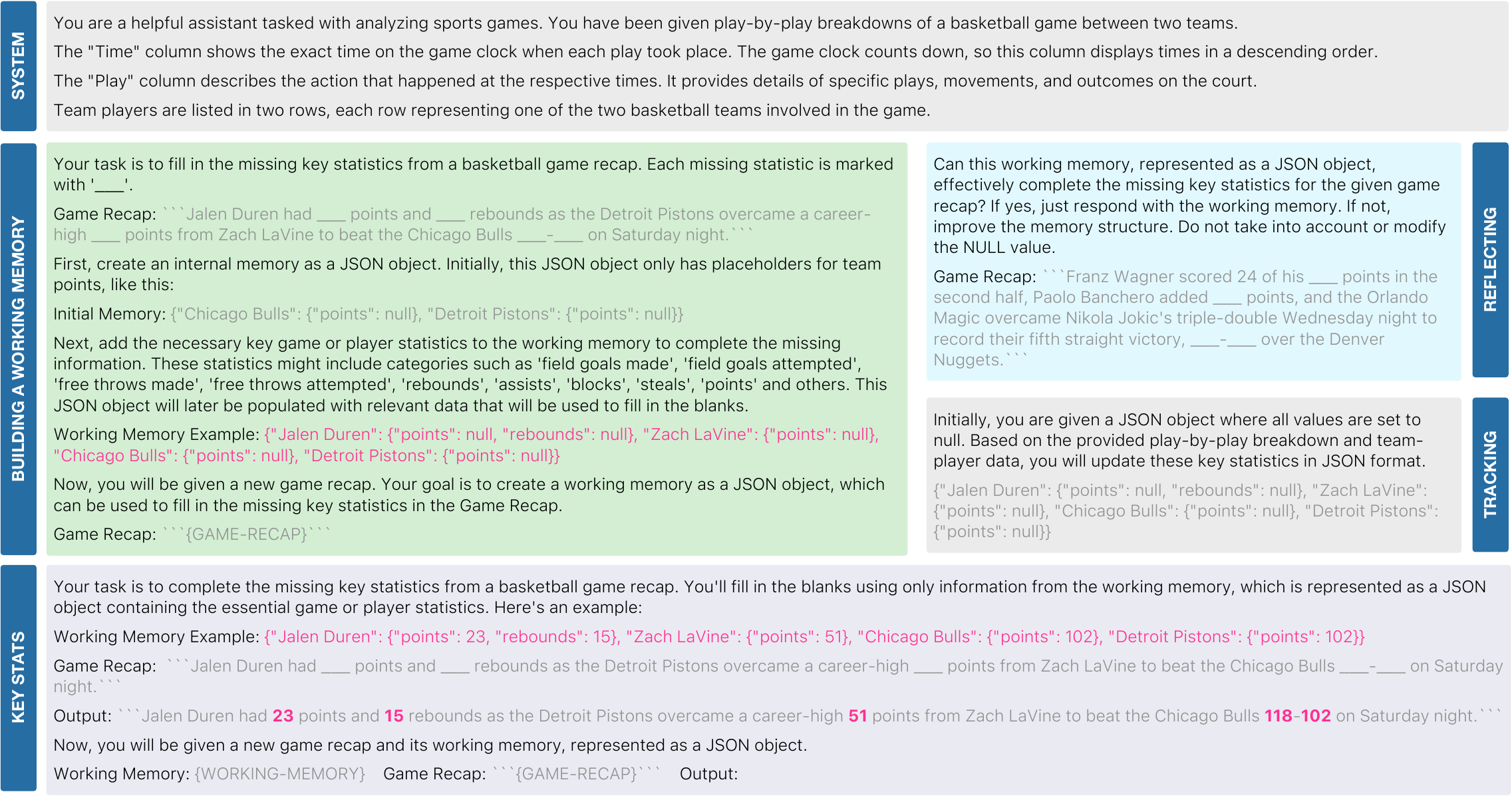}
\vspace{-0.2in}
\caption{An LLM fills in missing key statistics in game summaries through a three-step process. Initially, the LLM creates an internal JSON object as its memory. It then enriches this memory by adding necessary game or player statistics, where all values are set to null, and further reflects on whether this memory is sufficient to accomplish the task. Lastly, the LLM uses detailed play-by-play and team-player data to update the JSON object's values; it finally utilizes this updated memory to fill in the blanks in the game summary.
}
\label{fig:memory}
\vspace{-0.15in}
\end{figure*}

\section{The \texttt{SportsMetrics} Benchmark}
\label{sec:approach}

We collect NBA and NFL play-by-play data from ESPN.com. The descriptions capture the essence of each game. They are typically written by ESPN's sports journalists, who are experts in their respective sports. We reached out to ESPN as necessary to ensure adherence to their data policies. In Figure~\ref{fig:example-data}, we use ``\emph{time}'' to indicate the exact moment of each action on the game clock, while ``\emph{play}'' details the actions occurring at those times. Scoring actions, which change the game's score, are identified but not disclosed to LLMs during our experiments, as are team points. Additionally, we collect data on players' team affiliations and the game's box scores for our analysis.

Our task requires LLMs to track key stats across thousands of play-by-play records, which is a non-trivial effort. An ideal LLM needs to associate each action with the right player and their team in order to calculate team-level statistics. It must also monitor multiple key statistics simultaneously, such as field goals, free throws, rebounds, assists, blocks, steals, personal fouls, and turnovers in a basketball game. We believe an LLM's ability to summarize key details and fill in the missing statistics in game summaries demonstrates its capabilities in data fusion and numerical reasoning.

We need a comprehensive scoring metric to evaluate LLMs' ability to monitor key game statistics. Simply reporting individual metrics such as team points, rebounds, assists, and blocks for each team is inefficient and does not provide a holistic view of game analysis. To address this, we employ expert-developed team statistics formulas, as illustrated in Figure~\ref{fig:game-scores}. We adopt the NBA's ``\emph{\textbf{Game Score}}'' by John Hollinger, originally for player evaluation, to measure a team's overall effectiveness in basketball. It considers both positive (points, rebounds, assists) and negative (missed shots, turnovers) factors. For American football, we apply NCAA's ``\emph{\textbf{Passing Efficiency}}'' formula, as the NFL rule is more complex. 
In the following sections, we evaluate LLMs under different adversarial scenarios to assess their robustness.

\subsection{Long-Form Game Narratives}
\label{sec:long-input}

We begin by examining LLMs' ability to reason over long contexts. For example, Liu et al.~\shortcite{liu2023lost} introduced two tasks, multi-document QA and key-value retrieval, which require the model to identify relevant information within long contexts. They found that LLMs' performance significantly deteriorates when they have to access relevant information in the middle of long contexts. Our study goes a step further, requiring LLMs to not only identify relevant actions but also accurately track statistics throughout long-form game narratives. 

In this task, each LLM is provided detailed play-by-play descriptions of a sports game, including timestamps and specific actions. The players' team affiliations are listed in two rows, representing each team. The LLM's task is to use the play-by-plays to update key game statistics within a JSON object, initially filled with null values. For long-context LLMs such as GPT-4 Turbo, Claude 2.1, and Gemini Pro~\cite{geminiteam2023gemini}, we provide the entire game's data at once for processing. For LLMs with 4k or 8k tokens context, we break the game down into four quarters. The LLM gathers statistics quarter by quarter. It generates a JSON object that holds values from each quarter. These are then added up to derive game-level statistics.

\begin{table*}
\setlength{\tabcolsep}{8pt}
\renewcommand{\arraystretch}{1}
\centering
\begin{tabular}[t]{lrrllr}
\textbf{Model} & \textbf{Release Date} & \textbf{Context Len} & \textbf{Input}\;\;\; & \textbf{Output} & \textbf{Organization} \\
\toprule
Claude-2.1 & 11.21.2023 & 200,000 & \$.008 & \$.024 & Anthropic \\
GPT-4-1106-preview & 11.06.2023 & 128,000 & \$.01 & \$.03 & OpenAI \\
Gemini-Pro & 12.06.2023 & 32,000 & \$.001 & \$.002 & Deepmind\\
GPT-3.5-Turbo-1106 & 11.06.2023 & 16,385 & \$.001 & \$.002 & OpenAI\\
Mistral-7B-Instruct-v0.1 & 09.27.2023 & 8,000 & --- & --- & Mistral\\
GPT-3.5-Turbo-0613 & 06.13.2023 & 4,096 & \$.0015 & \$.0015 & OpenAI \\
Llama-2-13B-Chat & 07.18.2023 & 4,096 & --- & --- & Meta \\
\bottomrule
\end{tabular}
\vspace{-0.05in}
\caption{LLMs used in this study. Prices are per 1,000 tokens. Llama-2 and Mistral-7B are free and open-source.
}
\label{tab:llms}
\vspace{-0.1in}
\end{table*}

We use comprehensive, expert-devised formulas to evaluate LLMs in tracking game statistics.
For NBA games, we monitor 11 key statistics: \emph{team points}, \emph{field goals made}, \emph{field goals attempted}, \emph{free throws made}, \emph{free throws attempted}, \emph{offensive rebounds}, \emph{defensive rebounds}, \emph{steals}, \emph{assists}, \emph{blocks}, and \emph{personal fouls}.\footnote{We exclude \emph{turnovers} from tracking due to limitations in the data. Play-by-play descriptions may not capture every turnover, making it difficult for the model to track them accurately. When necessary, we rely on the ground-truth Turnover count from the box score to calculate the Game Score.} Moreover, we calculate `Game Score' to measure a team's overall effectiveness in basketball. For NFL games, we track \emph{passing yards}, \emph{touchdowns}, \emph{interceptions}, and \emph{pass completions} and \emph{attempts}. These additional stats allow for the computation of `Passing Efficiency.'

\subsection{The Impact of Changing Game Rules}
\label{sec:changing-rules}

It is important to understand LLMs' ability to make decisions under changing world rules. LLMs possess extensive knowledge from pretraining on the Internet, books, and other texts. This knowledge, held in their parametric memory, might not always align with the external evidence given to the model. Therefore, LLMs need to adjust to changing rules. Xie et al.~\shortcite{xie2023adaptive} highlight the importance of knowing when to trust a model's own knowledge. Meng et al.~\shortcite{meng2023massediting} explored finetuning LLMs to alter specific knowledge, but such changes are often irreversible. Here, we propose two tasks to evaluate LLMs' abilities in adapting to new game rules.

\paragraph{New Scoring Rules} We examine the impact of changing game rules on final scores. For basketball, scoring events such as free throws, three-pointers, field goals, vary from 1 to 3 points. We ask LLMs to maintain these scoring events but under a new rule where each action is worth only 1 point. This contradicts LLMs' existing knowledge, challenging them to recalibrate game scores accordingly. Ground-truth scores under this rule are obtained by counting the total number of scoring actions to determine each team's total points.

\paragraph{Player Swapping}  We randomly swapped player team affiliations in the table without changing the game's play-by-play records, as illustrated in Figure~\ref{fig:tasks}. Ground-truth team scores for this task are calculated by summing individual player scores under their new affiliations. This task allows us to vary the degree of conflict between the model's existing knowledge and the provided evidence. Swapping more players increases the task's difficulty.

\begin{table*}
\centering
\setlength{\tabcolsep}{5.3pt}
\renewcommand{\arraystretch}{1}
\begin{tabular}{llrrrrr}
& \textbf{System} & $\Delta$\textbf{GScore} & $\Delta$\textbf{Points} & $\Delta$\textbf{NewRule} & $\Delta$\textbf{Swap} & $\Delta$\textbf{Shuffle}\\
\toprule

\textbf{Long-Context} & GPT-3.5-Turbo-1106 & 33.50& \textbf{9.45} & \textbf{14.10} & \textbf{13.53} & \textbf{9.89} \\ 
(16k+ Tokens) & Gemini-Pro & \textbf{32.30}& 17.62 & 25.99  & 17.78 & 14.85 \\
& GPT-4-1106-preview & 51.97 & 25.17 & 14.55 & 39.91 & 49.57\\
& Claude-2.1 & 55.16& 21.73 & 22.28  & 17.12 & 31.11 \\
\midrule
\textbf{Standard} & GPT-3.5-Turbo-0613 & 114.28& 94.34 & {18.22} & 88.25 & 89.11 \\
(4k to 8k Tokens) & Mistral-7B-Instuct& 123.49 & 73.53 & 26.79 & 70.69  & 103.24 \\
& Llama-2-13B-Chat& 110.69 & 70.77 & 83.04 & 53.98 & 81.09 \\
\bottomrule
\end{tabular}
\caption{Average absolute difference between model predictions and the actual scores on NBA data for tracking a team's total points (\textbf{Points}) and all key game statistics (\textbf{GScore}). Moreover, we evaluate LLMs' performance in three adversarial scenarios: $\Delta$\textbf{NewRule}, $\Delta$\textbf{Swap} and $\Delta$\textbf{Shuffle}.}
\label{table:nba-results}
\end{table*}

\subsection{Robustness Against Noise}
\label{sec:noise-robustness}

\paragraph{Shuffling Play-by-Plays} We present an adversarial challenge where we shuffle basketball game play-by-play descriptions and then ask LLMs to track the total points of each team. We choose basketball because adjacent actions in this context do not show strong causal relationships. Changing the sequence of scoring actions does not affect the teams' total points. We anticipate that long-context LLMs will produce consistent or similar final game scores. To avoid confusing the model, we maintain the original order of timestamps.

We can also adjust the frequency of scoring plays in a game, making it more or less challenging for LLMs to process the narrative. By choosing a probability $p$ from a set of values \{-50\%, -20\%, 0, +20\%, +50\%\}, we can either duplicate non-scoring plays (thereby decreasing scoring play density and extending the game narrative) or remove them (increasing scoring play density). Further, to test the LLM's inherent knowledge, we randomly select players from each team in NFL games and assign them new names, such as characters from science fictions. This approach evaluates the model's ability to adapt to changes in player identities. These alterations do not introduce new players or change the total points scored in the game; it simply varies the narrative's complexity.

\begin{figure}
\centering
\includegraphics[width=3in]{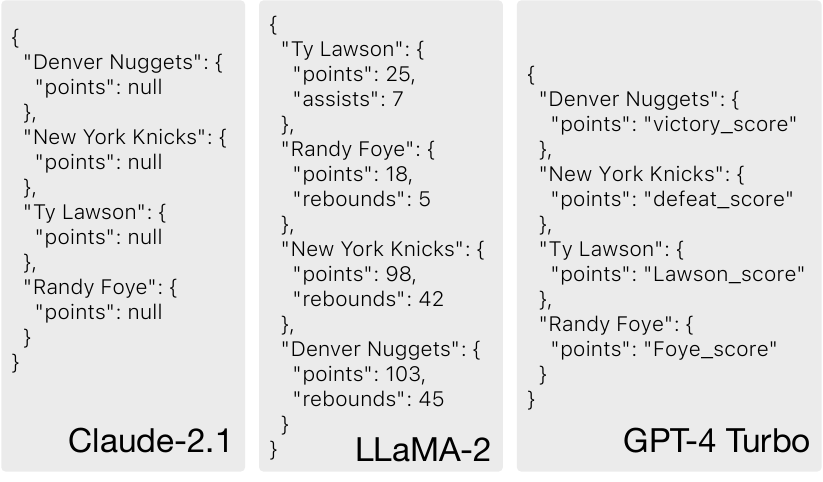}
\vspace{-0.1in}
\caption{Effective working memory is key in this task. 
The variance in memory structure arises because we allowed each LLM to generate its JSON object as working memory, without enforcing a uniform schema. This step allows us to explore how each model organizes its memory to complete the task.
Note that Claude's `null' values represent an initial state rather than an inability to aggregate information. 
}
\label{fig:json-obj}
\vspace{-0.1in}
\end{figure}

\subsection{Planning for Complex Data Queries}
\label{sec:memory}

In this task, LLMs fill in missing key statistics from game summaries (e.g., from ESPN). The process unfolds in three steps, illustrated in Figure~\ref{fig:memory}. First, the LLM creates an internal JSON object memory, initially with placeholders for team points. Next, it enriches this memory by adding crucial game or player statistics. During a self-reflection phase, the LLM evaluates if its JSON memory can accurately complete the missing statistics for the given game recap. If it can, it responds with this memory; if not, it further refines the memory structure. Finally, using the detailed play-by-play and team-player data, the LLM updates the key statistics in the JSON format, then uses this information to fill in the blanks in the game summary. Figure~\ref{fig:json-obj} illustrates various LLM attempts building a memory. 

Our task is inspired by several studies on LLM planning. Unlike LLM+P which uses the Planning Domain Definition Language (PDDL) for problem-solving~\cite{liu2023llmp}, we simplify the process by requiring only a valid JSON object for working memory. Relevant studies such as Reflexion~\cite{shinn2023reflexion}, ReAct~\cite{yao2023react}, and Tree-of-thought~\cite{yao2023tree} have also influenced our approach. Sumers et al.~\shortcite{sumers2023cognitive} have developed a framework for integrating planning into LLM agents. Prior studies have focused on ALFWorld's interactive TextWorld environments. Our method are focused on sports, which involves masking key statistics in game recaps by sports journalists, then converting them into task data points for LLMs. We assess LLMs by their accuracy in filling in missing key statistics from game summaries.

\section{Experiments}
\label{sec:results}

\begin{table*}
\centering
\setlength{\tabcolsep}{5pt}
\renewcommand{\arraystretch}{1}
\begin{tabular}{llrrrrrr}
& \textbf{System} & $\Delta$\textbf{Yards} & \;$\Delta$\textbf{ATT} & $\Delta$\textbf{COMP} & \;$\Delta$\textbf{TD} & $\Delta$\textbf{INT} & \quad$\Delta$\textbf{PE} \\ 
\toprule
\textbf{Long-Context} & GPT-4-1106-Preview & \textbf{34.77} & \textbf{4.44} & \textbf{2.96} & \textbf{0.17} & \textbf{0.13} & \textbf{14.33}\\
(16k+ Tokens) & Claude-2.1 & 52.53 & 5.43 & 3.75 & 0.29 & 0.22 & {17.53} \\
& GPT-3.5-Turbo-1106 & 64.87 & 7.80 & 4.73  & 0.49 & 0.30 & {18.43}\\
& Gemini-Pro & 85.14  & 12.68 & 6.87 & 0.83 & 0.52 & {26.17} \\
\midrule
\textbf{Standard} & GPT-3.5-0613 & 105.68 & 24.11 & 15.80 & 1.09 & 0.60 & \textbf{89.56} \\
(4k to 8k Tokens) & Llama-2-13B-Chat& 244.48 & 22.37 & 19.66 & 1.47 & 1.03 & 191.76 \\
& Mistral-7B-Instuct& 119.31 & 17.64 & 9.05 & 1.23 & 0.69 & 202.07  \\
\bottomrule
\end{tabular} 
\vspace{-0.1in}
\caption{
Discrepancies between model predictions and actual scores on NFL stats, including yards (\textbf{Yards}), attempts (\textbf{ATT}), completions (\textbf{COMP}), touchdowns (\textbf{TD}), interceptions (\textbf{INT}) and passing efficiency (\textbf{PE}). 
}
\label{table:nfl-stats-results}
\end{table*}

We evaluate various LLMs in our \texttt{SportsMetrics} benchmark. These models are listed in Table~\ref{tab:llms} and split into two categories: long-context LLMs, capable of processing over 16k tokens, and standard LLMs, handling 4k to 8k tokens. Our evaluation focuses on their ability to accurately track a team's total points (\textbf{Points}) and all key game statistics (\textbf{GameScore}). We measure the average absolute difference (deviation) between the models' predictions and the actual box scores, denoted as $\Delta$Points and $\Delta$GScore, respectively.\footnote{$\Delta$GScore consistently shows higher values compared to $\Delta$Points because it goes beyond counting a team's points. It offers a full game analysis by requiring the LLM to consolidate key statistics such as points, rebounds, steals, assists and more into an overall score. Considering only team points is insufficient, especially in sports like soccer where scoring is rare. When necessary, we can convert GameScore to points by zeroing out other stats.}

Our dataset comprises 28,492 NBA games and 5,867 NFL games spanning two decades from 2002 to 2023, available through ESPN's archives. We randomly selected 100 games from each sport for our test set. On average, NBA games contain 466 plays and NFL games 173 plays. An average NBA game includes 6,229 tokens, while an NFL game has 6,166 tokens, with maximum lengths reaching 7,322 and 7,659 tokens, respectively.

LLMs' ability to integrate information is tested under three adversarial scenarios: (a) `\textbf{NewRule},' which assigns every scoring action just one point, regardless of the move, (b) `\textbf{Swap}' which randomly selects two players from each team to swap their affiliations in the team-player table, (c) `\textbf{Shuffle},' which duplicates any non-scoring action with a 20\% chance ($p$=0.2) before shuffling the play-by-plays. We assess LLMs' performance in these scenarios and report the deviation of predicted team points from actual scores as $\Delta$NewRule, $\Delta$Swap and $\Delta$Shuffle.

In Table~\ref{table:nba-results}, we present our findings from the NBA section of our dataset. With $\Delta$ representing the gap between predictions and actual scores, smaller values are preferable. We find that long-context LLMs significantly outperform standard LLMs across all tasks. GPT-3.5-Turbo-1106 leads in performance in every task except for $\Delta$GScore, where Gemini-Pro has a slight edge. Long-context models have been released recently in late 2023. These results demonstrate their remarkable ability in identifying relevant actions from game play-by-plays, attributing each action to the right player and team, and aggregating numerical data to compute final team points and GameScore.
This requires a level of numerical reasoning that humans are adept at but it is still new territory for LLMs.

\begin{figure}
\centering
\includegraphics[width=3in,height=2.35in]{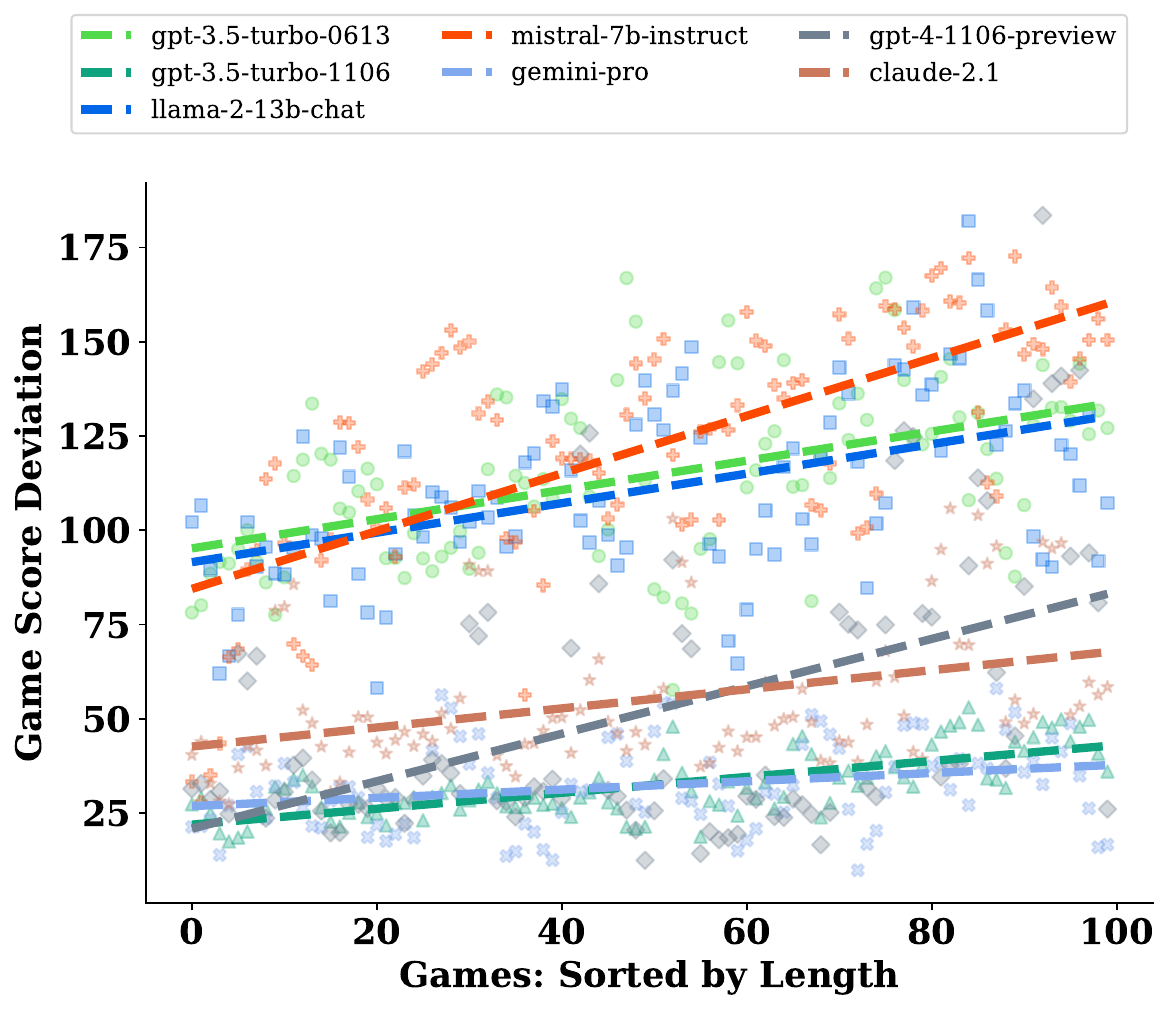}
\vspace{-0.2in}
\caption{We organize games based on the length of their play-by-plays, with the x-axis showing the games and the y-axis the deviation scores; lower scores indicate better performance. GPT-3.5-Turbo-1106 and Gemini-Pro stand out here, maintaining nearly flat curves.
}
\label{fig:episodic-memory-results}
\vspace{-0.1in}
\end{figure}

In Figure~\ref{fig:episodic-memory-results}, we organize games based on the length (number of tokens) of their play-by-play descriptions, with the x-axis showing the games and the y-axis the deviation scores from various LLMs, where lower scores indicate better performance. We perform a regression analysis to demonstrate each LLM's trend in handling games of increasing length. GPT-3.5-Turbo-1106 and Gemini-Pro stand out, maintaining nearly flat curves, which corresponds with their superior performance as shown in Table~\ref{table:nba-results}. By contrast, GPT-4-1106-Preview does well in shorter games but face difficulties in aggregating key statistics for longer games. Additionally, 79\% its returned JSON objects contain zeros or null values, contributing to its unsatisfying performance on this task.

We note that basketball teams typically score between 100 to 120 points. Our findings show that the smallest prediction gap for $\Delta$Points is 9.45, while the largest can exceed 100. This indicates the difficulty in accurately tracking key game statistics over long contexts, as \emph{standard LLMs can produce predictions significantly off from actual scores due to hallucinations.} Among the three adversarial scenarios, the New Rule is relatively simpler as it requires LLMs to assign one point to every scoring action, focusing on counting these actions instead of distinguishing between types (3-pointers vs. free throws) and adding them up for a team's score. In this scenario, Llama-2-13B-Chat scores lower than all other LLMs.

\begin{figure*}
\centering
\includegraphics[width=6.2in]{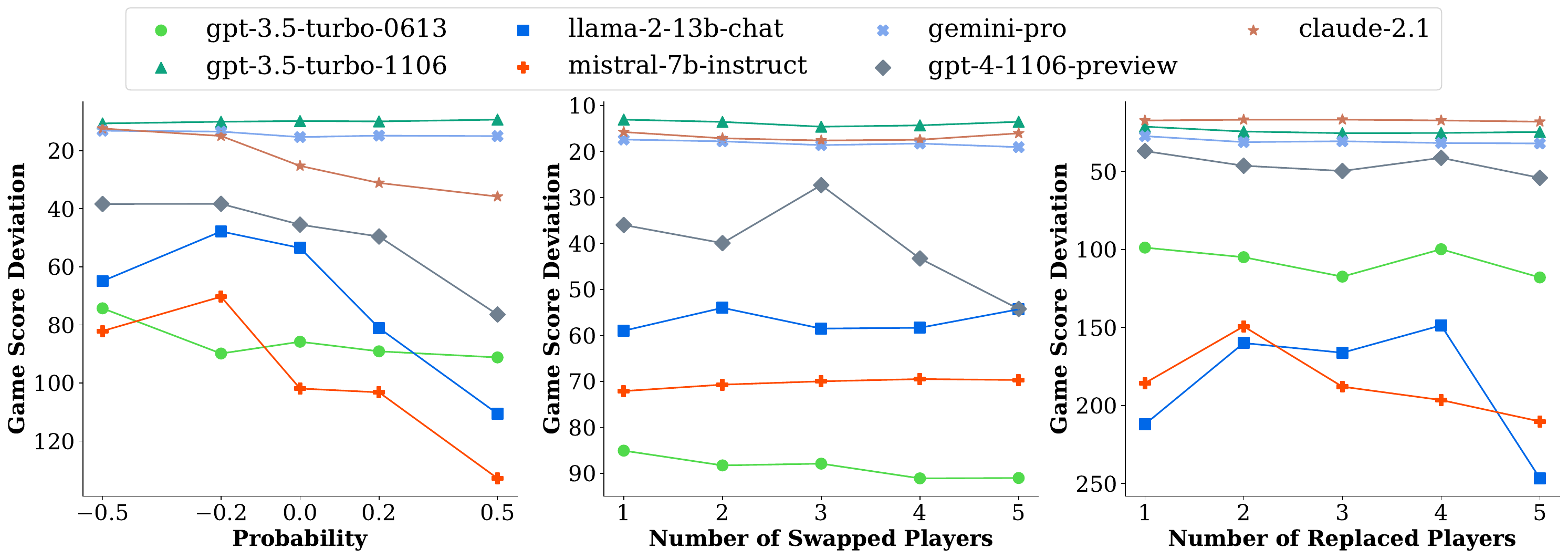}
\vspace{-0.05in}
\caption{
(\textsc{Left}) We adjust the difficulty of identifying scoring events by either removing or duplicating non-scoring events. Moreover, we randomly swapped $n$ players' affiliations in the team-player table (\textsc{Middle}) and replaced $n$ players' names with science fiction characters (\textsc{Right}), all without changing the play-by-play texts.
}
\label{fig:curves}
\vspace{-0.1in}
\end{figure*}

In Table~\ref{table:nfl-stats-results}, we present NFL data findings. American football's play-by-plays have demonstrated a sequential nature, we cannot apply tests like New Rule, Swap, or Shuffle as with basketball games. Instead, we measure how model predictions deviate from actual scores on key game statistics, including yards (\textbf{$\Delta$Yards}), attempts (\textbf{$\Delta$ATT}), completions (\textbf{$\Delta$COMP}), touchdowns (\textbf{$\Delta$TD}), and interceptions (\textbf{$\Delta$INT}). We also combine them into Passing Efficiency (\textbf{$\Delta$PE}) for a holistic game analysis. Our results suggest that long-context LLMs greatly surpass standard models, with GPT-4-1106-Preview taking the lead, followed by Claude-2.1 and GPT-3.5-Turbo-1106. 

Particularly, passing yards are vital in the NFL games, often leading to scoring opportunities like touchdowns and field goals. On average, NFL teams average 200 to 250 passing yards per game. We find that the top model, GPT-4-1106-Preview, exhibits a 34.77-yard discrepancy in passing yards prediction, while the open-source Llama-2-13B-Chat lags significantly in comparison. This highlights the difficulty of tracking passing yards, a task even more challenging than summarizing basketball points, with most models struggling to accurately aggregate such data.

Our results suggest that the difference in performance between GPT-3.5-Turbo-1106 and GPT-4 across basketball and football games stems from the scoring frequency in each sport. Basketball's frequent scoring presents a challenge for GPT-4-1106-Preview to track all actions, while football, with less frequent scoring, is somewhat easier for the model to track. GPT-4-1106-Preview is optimized for handling extremely long contexts and it is less accurate in tracking frequent scoring. This distinct characteristic accounts for the varied performance of both models.

\begin{figure}
\centering
\includegraphics[width=3in, height=2.2in]{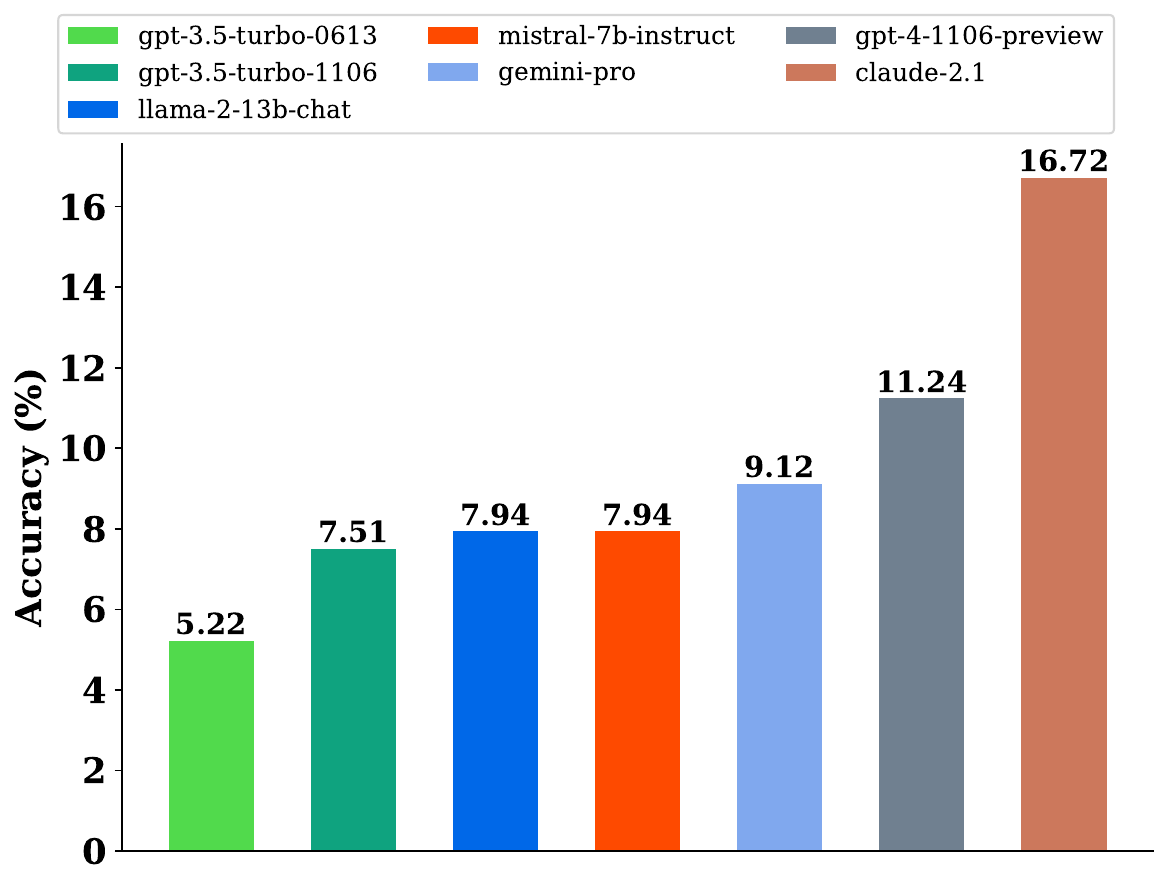}
\vspace{-0.2in}
\caption{Accuracy of various LLMs in filling missing key statistics from basketball game recaps. Claude-2.1 shows strong performance, while Mistral-7B-Instruct achieves the highest accuracy among standard LLMs. }
\label{fig:mem-results}
\vspace{-0.15in}
\end{figure}

In Figure~\ref{fig:curves}, we test LLMs' robustness against adversarial conditions. In the left subfigure, we vary the difficulty of identifying scoring events by either dropping or duplicating non-scoring events. E.g., at probability $p$=-0.5, we eliminate any non-scoring event with a 50\% chance; at $p$=0.2, we duplicate any non-scoring event with a 20\% chance, before shuffling the entire game description. The y-axis measures the deviation from the actual box score, with smaller values indicating better model performance. We observe that \emph{GPT-3.5-Turbo-1106 and Gemini-Pro perform the best, whose curves are quite flat, indicating their robustness to a varying level of noise in the play-by-plays.} Overall, LLMs perform well when non-scoring events are removed, yet their performance drops as more non-scoring events are added, akin to searching for a needle in a larger haystack.

Further, we randomly swapped $n$ players' affiliations in the team-player table and replaced $n$ players' names with science fiction characters, all without changing the play-by-play texts. Our findings are shown in the middle and right subfigures. We find that Claude-2.1, Gemini-Pro, and GPT-3.5-Turbo-1106 are the top performers. Interestingly, \emph{renaming players significantly decreases all models' performance. This suggests LLMs may use familiar basketball player names from their pretraining to guess team scores, rather than analyzing the actual play-by-plays.} GPT-4-1106-Preview is the least adaptable to these adversarial conditions among the long-context LLMs. We also observe a notable performance disparity exists between open-source and proprietary LLMs.

We assess the accuracy of various LLMs in completing missing key statistics from basketball game recaps. The types of missing data include a player's total points, team scores, assists, rebounds, and other stats. An LLM must understand the recap's context to precisely estimate the missing statistic. To do this, LLMs create a JSON object as its working memory. They then calculate the needed statistics using play-by-play and team-player data and use this memory object to fill in the blanks.

Figure~\ref{fig:mem-results} presents the results of this task. Claude-2.1 shows strong performance, while Mistral-7B-Instruct achieves the highest accuracy among standard LLMs. This task requires that LLMs possess strong instruction-following capabilities to build an effective working memory. Figure~\ref{fig:json-obj} provides sample working memories from various LLMs. Although complex structures are possible, they increase the risk of errors when populating values. Models such as GPT-4-1106-Preview and Llama-2-13B-Chat face difficulties in creating a working memory. They hallucinate field values or fail to accurately fill fields with aggregated values from play-by-play data. By contrast, Claude-2.1's memory structure is the best in terms of efficiency, focusing on essential game statistics. Our task crucially evaluates LLMs' memory management skills when handling complex data queries.

\section{Conclusion}
\label{sec:conclusion}

We introduce \texttt{SportsMetrics}, a novel benchmark designed to evaluate LLMs in sports data analytics. It assess LLMs' numerical reasoning and fusion abilities through challenges such as new game rules, lengthy descriptions, scrambled narratives and key stats analysis in game summaries. \texttt{SportsMetrics} highlights LLMs' potential in fields such as multiplayer gaming and collaborative workspaces.

\section{Limitations}
\label{sec:limitations}

Our research focuses on NBA and NFL games, which are major sports with rich datasets. We are interested in exploring the generalizability of our findings to other sports. For example, soccer and cricket have distinct play styles and rules, which might challenge LLMs in unique ways. Our study has explored multiple adversarial scenarios, such as new game rules and scrambled game narratives. Such drastic changes might be uncommon in real-world conditions, and the models' ability to handle these scenarios might not translate to improved performance in other analytical tasks. Finally, our scoring system's effectiveness in assessing LLMs' numerical reasoning capabilities in different contexts, such as multiplayer online gaming or collaborative workspaces, remains to be validated. This study explores LLMs' potential in sports analytics. It is important to recognize these limitations when applying our findings to broader contexts.

\section*{Acknowledgements}
We are grateful to the reviewers for their insightful feedback, which has helped enhance the quality of our paper. This research has been partially supported by the NSF CAREER award, \#2303655.


\bibliography{custom,anthology}

\end{document}